# Explanations of Black-Box Model Predictions by Contextual Importance and Utility

Sule Anjomshoae, Kary Främling and Amro Najjar

Department of Computing Science, Umeå University, Umeå, Sweden
{sule.anjomshoae, kary.framling, amro.najjar}@umu.se

**Abstract.** The significant advances in autonomous systems together with an immensely wider application domain have increased the need for trustable intelligent systems. Explainable artificial intelligence is gaining considerable attention among researchers and developers to address this requirement. Although there is an increasing number of works on interpretable and transparent machine learning algorithms, they are mostly intended for the technical users. Explanations for the end-user have been neglected in many usable and practical applications. In this work, we present the Contextual Importance (CI) and Contextual Utility (CU) concepts to extract explanations that are easily understandable by experts as well as novice users. This method explains the prediction results without transforming the model into an interpretable one. We present an example of providing explanations for linear and non-linear models to demonstrate the generalizability of the method. CI and CU are numerical values that can be represented to the user in visuals and natural language form to justify actions and explain reasoning for individual instances, situations, and contexts. We show the utility of explanations in car selection example and Iris flower classification by presenting complete (i.e. the causes of an individual prediction) and contrastive explanation (i.e. contrasting instance against the instance of interest). The experimental results show the feasibility and validity of the provided explanation methods.

**Keywords:** Explainable AI, black-box models, contextual importance, contextual utility, contrastive explanations.

## 1 Introduction

Intelligent systems are widely used for decision support across a broad range of industrial systems and service domains. A central issue that compromise the adoption of intelligent systems is the lack of explanations for the actions taken by them. This is a growing concern for effective human-system interaction. Explanations are particularly essential for intelligent systems in medical diagnosis, safety-critical industry, and automotive applications as it raises trust and transparency in the system. Explanations also help users to evaluate the accuracy of the system's predictions [1]. Due to a growing need for intelligent systems' explanations, the field of eXplainable Artificial

Intelligence (XAI) is receiving a considerable amount of attention among developers and researchers [2].

While generating explanations have been investigated in early years of expert systems, intelligent systems today have become immensely complex and rapidly evolving in new application areas. As a result, generating explanation for such systems is more challenging and intriguing than ever before [3, 4]. This is particularly relevant and important in intelligent systems that have more autonomy in decision making. Nonetheless, as important as it is, existing works are mainly focusing on either creating mathematically interpretable models or converting black-box algorithms into simpler models. In general, these explanations are suitable for expert users to evaluate the correctness of a model and are often hard to interpret by novice users [5, 6]. There is a need for systematic methods that considers the end user requirements in generating explanations.

In this work, we present the Contextual Importance (CI) and Contextual Utility (CU) methods which explain prediction results in a way that both expert and novice users can understand. The CI and CU are numerical values which can be represented as visuals and natural language form to present explanations for individual instances [7]. Several studies suggested modeling explanation facilities based on practically relevant theoretical concepts such as contrastive justifications to produce human understandable explanations along with the complete explanations [8]. Complete explanations present the list of causes of an individual prediction, while contrastive explanations justify why a certain prediction was made instead of another [9]. In this paper, we aim at providing complete explanations as well as the contrastive explanations using CI and CU methods for black-box models. This approach generally can be used with both linear and non-linear learning models. We demonstrate an example of car selection problem (e.g. linear regression) and classification problem (e.g. neural network) to extract explanations for individual instances.

The rest of the paper is organized as follows: Section 2 discusses the relevant background study. Section 3 reviews the state of the art for generating explanation. Section 4 explains the contextual importance and utility method. Section 5 presents the explanation results for the regression and classification example. Section 6 discusses the results and, Section 7 concludes the paper.

## 2  Background

Explanations were initially discussed in rule-based expert systems to support developers for system debugging. Shortliffe's work is probably the first to provide explanation in a medical expert system [10]. Since then, providing explanations for intelligent systems' decisions and actions has been a concern for researchers, developers and the users. Earlier attempts were limited to traces, and a line of reasoning explanations that are used by the decision support system. However, this type of explanations could only be applied in rule-based systems and required knowledge of decision design [11]. These systems were also unable to justify the rationale behind a decision.

Swartout's framework was one of the first study that emphasized the significance of justifications along with explanations [12]. Early examples proposed justifying the outcomes through drilling-down into the rationale behind each step taken by the system. One approach to produce such explanation was storing the justifications as canned text for all the possible questions that can be inquired [13]. However, this approach had several drawbacks such as maintaining the consistency between the model and the explanations, and predicting all the possible questions that the system might encounter.

The decision theory was proposed to provide justifications for the system's decision. Langlotz suggested decision trees to capture uncertainties and balance between different variables [14]. Klein developed explanation strategies to justify value-based preferences in the context of intelligent systems [15]. However, these explanations required knowledge of the domain in which the system will be used [16]. This kind of explanation were less commonly used, due to the difficulties in generating such explanations that satisfies the needs of the end-users [11].

Expert systems that are built based on probabilistic decision-making systems such as Bayesian networks required the explanations even more due to their internal logic is unpredictable [17]. Comprehensive explanations of probabilistic reasoning are therefore studied in a variety of applications to increase the acceptance of expert systems [18]. Explanation methods in Bayesian networks have been inadequate to constitute a standard method which is suitable for systems with similar reasoning techniques.

Previous explanation studies within expert systems are mostly based on strategies that rely on knowledge base and rule extraction. However, these rule-based systems and other symbolic methods perform poorly in many explanation tasks. The number of rules tends to grow extremely high, while the explanations produced are limited to showing the applicable rules for the current input values. The Contextual Importance and Utility (CIU) method was proposed to address these explanation problems earlier [7]. This method explains the results directly without transforming the knowledge. The details of this work are discussed in Section 4.

## 3    State of the Art

Machine learning algorithms are the heart of many intelligent decision support systems in finance, medical diagnosis, and manufacturing domains. Because some of these systems are considered as black-box (i.e. hiding inner-workings), researchers have been focusing on integrating explanation facilities to enhance the utility of these systems [19]. Recent works define interpretability particular to their explanation problems. Generally these methods are categorized into two broad subject-matter namely, model-specific and model-agnostic methods. The former one typically refers to inherently interpretable models which provide a solution for a predefined problem. The latter provides generic framework for interpretability which is adaptable to different models.

In general, model-specific methods are limited to certain learning models. Some intrinsically interpretable models are sparse linear models [20, 21], discretization methods such as decision trees and association rule lists [22, 23], and Bayesian rule lists [24]. Other approaches include instance-based models [25] and mind-the-gap model [26] focus on creating sparse models through feature selection to optimize interpretability. Nevertheless, linear models are not that competent at predictive tasks, because the relationships that can be learned are constrained and the complexity of the problem is overgeneralized. Even though they provide insight into why certain predictions are made, they enforce restrictions on the model, features, and the expertise of the users.

Several model-agnostic frameworks have been recently proposed as an alternative to interpretable models. Some methods suggest measuring the effect of an individual feature on a prediction result by perturbing inputs and seeing how the result changes [27, 28]. The effects are then visualized to explain the main contributors for a prediction and to compare the effect of the feature in different models. Ribeiro et al. [29] introduce Locally Interpretable Model Explanation (LIME) which aims to explain an instance by approximating it locally with an interpretable model. The LIME method implements this by sampling around the instance of interest until they arrive at a linear approximation of the global decision function. The main disadvantage of this method is that data points are sampled without considering the correlation between features. This can create irrelevant data points which can lead to false explanations. An alternative method is Shapley values where the prediction is fairly distributed among the features based on how each feature contributes to the prediction value. Although, this method generates complete and contrastive explanations, it is computationally expensive. In general, model-agnostic methods are more flexible than model-specific ones. Nevertheless, the correctability of the explanations and incorporating user feed-back in explanation system are still open research issues [30].

## 4     Contextual Importance and Contextual Utility

Contextual importance and utility were proposed as an approach for justifying recommendations made by black-box systems in Kary Främling's PhD thesis [31], which is presumably one of the earliest studies addressing the need to explain and justify specific recommendations or actions to end users. The method was proposed to explain preferences learned by neural networks in a multiple criteria decision making (MCDM) context [32]. The real-world decision-making case consisted in choosing a waste disposal site in the region of Rhône-Alpes, France, with 15 selection criteria and over 3000 potential sites to evaluate. A similar use case was implemented based on data available from Switzerland, as well as a car selection use case. In such use cases, it is crucial to be able to justify the recommendations of the decision support system also in ways that are understandable for the end-users, in this case including the inhabitants of the selected site(s).

Multiple approaches were used for building a suitable MCDM system, i.e. the well-known MCDM methods Analytic Hierarchy Process (AHP) [33] and ELECTRE [34]. A rule-based expert system was also developed. However, all these approaches suffer from the necessity to specify the parameters or rules of the different models, which needs to be based on a consensus between numerous experts, politicians and other stakeholders. Since such MCDM systems can always be criticized for being subjective, a machine learning approach that would learn the MCDM model in an "objective" way based on data from existing sites became interesting.

MCDM methods such as AHP are based on *weights* that express the *importance* of each input (the selection criteria) for the final decision. A notion of *utility* and *utility function* is used for expressing to what extent different values of the selection criteria are favorable (or not) for the decision. Such MCDM methods are linear in nature, which limits their mathematical expressiveness compared to neural networks, for instance. On the other hand, the weights and utilities give a certain transparency, or explainability, to the results of the system. The rationale behind Contextual Importance (CI) and Contextual Utility (CU) is to generalize these notions from linear models to non-linear models [7].

In practice, the importance of criteria and the usefulness of their values change according to the current context. In cold weather, the importance and utility of warm clothes increases compared to warm summer weather, whereas the importance of the sunscreen rating that might be used becomes small. This is the reason for choosing the word contextual to describe CI and CU. This approach generally can be used with both linear and non-linear learning models. It is based on explaining the model's predictions on individual importance and utility of each feature.

CI and CU are defined as:

$$CI = \frac{Cmax_x(C_i) - Cmin_x(C_i)}{absmax - absmin} \quad (1)$$

$$CU = \frac{y_{i,j} - Cmin_x(C_i)}{Cmax_x(C_i) - Cmin_x(C_i)} \quad (2)$$

where

- $C_i$ is the context studied (which defines the fixed input values of the model),
- $x$ is the input(s) for which CI and CU are calculated, so it may also be a vector,
- $y_{i,j}$ is the output value for the output $j$ studied when the inputs are those defined by $C_i$,
- $Cmax_x(C_i)$ and $Cmin_x(C_i)$ are the highest and the lowest output values observed by varying the value of the input(s) $x$,
- $absmax$ and $absmin$ specify the value range for the output j being studied.

CI corresponds to the fraction of output range covered by varying the value(s) of inputs *x* and the maximal output range. CU reflects the position of $y_{i,j}$ within the output range covered ($Cmax_x(C_i)$ - $Cmin_x(C_i)$). Each feature *x* with prediction $y_{i,j}$ has its own CI and CU values.

The estimation of $Cmax_x(C_i)$ and $Cmin_x(C_i)$ is a mathematical challenge, which can be approached in various ways. In this paper, we have used Monte-Carlo simula-tion, i.e. generating a "sufficient" number of input vectors with random values for the $x$ input(s). Obtaining completely accurate values for $Cmax_x(C_i)$ and $Cmin_x(C_i)$ would in principle require an infinite number of random values. However, for the needs of explainability, it is more relevant to obtain CI values that indicate the relative importance of inputs compared to each other. Regarding CU, it is not essential to obtain exact values neither for producing appropriate explanations. However, the estimation of $Cmax_x(C_i)$ and $Cmin_x(C_i)$ remains a matter of future studies. Gradient-based methods might be appropriate in order to keep the method model-agnostic. In [31], Normalized Radial Basis Function (RBF) networks were used, where it makes sense to assume that minimal and maximal output values will be produced at or close to the centroids of the RBF units. However, such methods are model-specific, i.e. specific to a certain type or family of black-box models.

CI and CU are numerical values that can be represented in both visual and textual form to present explanations for individual instances. CI and CU can also be calculated for more than one input or even for all inputs, which means that arbitrary higher-level concepts that are combinations of more than one inputs can be used in explanations. Since the concepts and vocabularies that are used for producing explanations are external to the black box, the vocabularies and visual explanations can be adapted depending on the user they are intended for. It is even possible to change the representation used in the explanations if it turns out that the currently used representation is not suitable for the user's understanding, which is what humans tend to do when another person does not seem to understand already tested explanation approaches. Fig. 1 illustrates how explanations are generated using contextual importance and utility method.

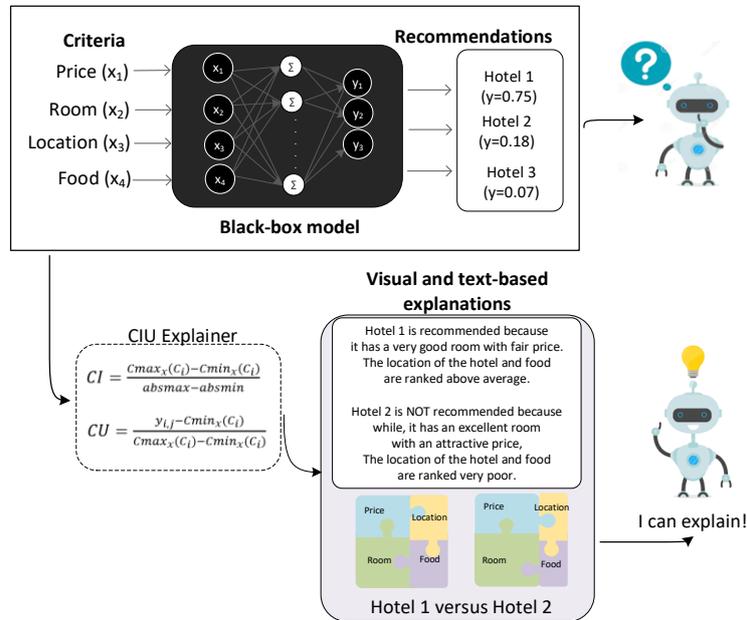

**Fig. 1.** Providing explanations for individual instances using CI and CU

Another important point is that humans usually ask for explanations of why a certain prediction was made instead of another. This gives more insight into what would be the case if the input had been different. Creating contrastive explanations and comparing the differences to another instance can often be more useful than the complete explanation alone for a particular prediction. Since the contextual importance and utility values can be produced for all possible input value combinations and outputs, it makes it possible to explain why a certain instance $C_i$ is preferable to another one, or why one class (output) is more probable than another. The algorithm used for producing complete and contrastive explanations is shown in Fig. 2.

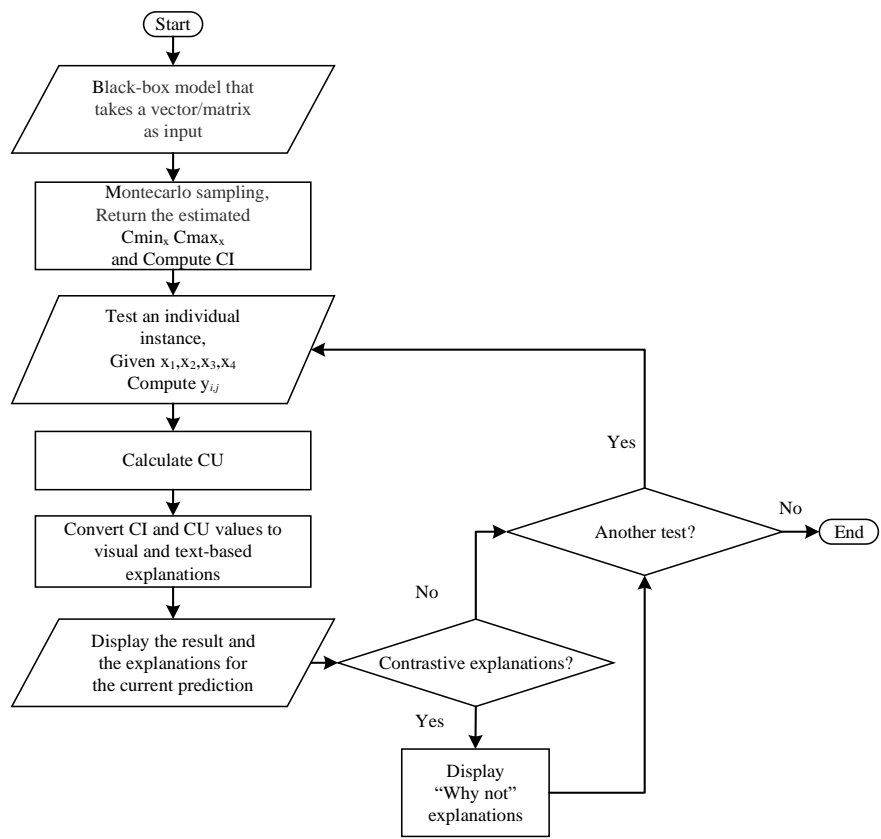

**Fig. 2.** Algorithm for generating complete and contrastive explanations using CIU method

## 5   Examples of CI and CU Method to Extract Explanation for Linear and Non-Linear Models

The explanation method presented here provides flexibility to explain any learning model that can be considered a "black-box". In this section, we present the examples of providing explanations for linear and non-linear models using contextual im-portance and utility method. Code explaining individual prediction for non-linear models is available at https://github.com/shulemsi/CIU.

### 5.1 Visual Explanations for Car Selection Using CI and CU Method

The result of explanations for a car selection problem using CI and CU method is presented. The dataset used in this example was initially created and utilized to learn the preference function by neural network in a multi-criteria decision-making problem [31]. Here, these samples are used to show how explanations can be generated for linear models. The dataset contains 113 samples with thirteen different characteristics of the car and their respective scores. Some of the characteristics are namely; price of the car, power, acceleration, speed, dimensions, chest, weight, and aesthetic. The linear relation between price and preference, and the corresponding CI and CU values are demonstrated in Fig. 3.

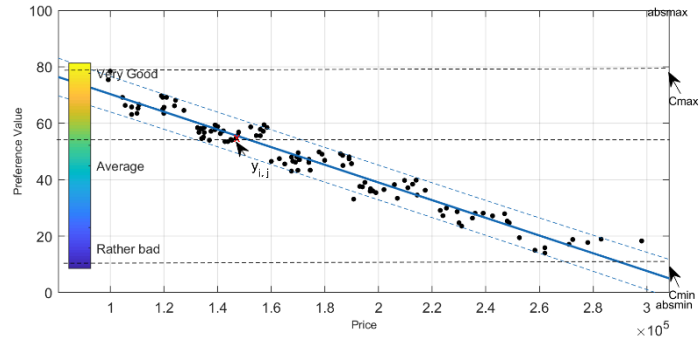

**Fig. 3.** The contextual importance and contextual utility values for price (selected car WW Passat GL).

The preference value is shown as a function of the price of the car, the red-cross ($y_{i,j}$) showing the current value for the selected car WW Passat GL. The color scale shows the limits for translating contextual utility values into words to generate text-based explanations. Contextual importance values are converted into words using the same kind of scale. Table 1 reveals $Cmax_x$, $Cmin_x$, CI and CU values of the price of the car and other key features including power, acceleration, and speed for the example car WW Passat GL.

**Table 1.** CI and CU values of the features price, power, acceleration, and speed for the selected car example WW Passat-GL

|       | Price | Power | Acceleration | Speed |
|-------|-------|-------|--------------|-------|
| **Cmin** | 13    | 14    | 13           | 10    |
| **Cmax** | 79    | 78    | 68           | 64    |
| **CI%**  | 66    | 64    | 55           | 54    |
| **CU**   | 0.67  | 0.15  | 0.30         | 0.25  |

The table shows that the price and power are the most important features for the selected car. Also, the highest utility value belongs to the price which means it is the most preferred feature of this car. The least preferred characteristic of this car is the power which has the lowest utility value. The CI and CU values led the following visual and text-based explanations as shown in Fig. 4.

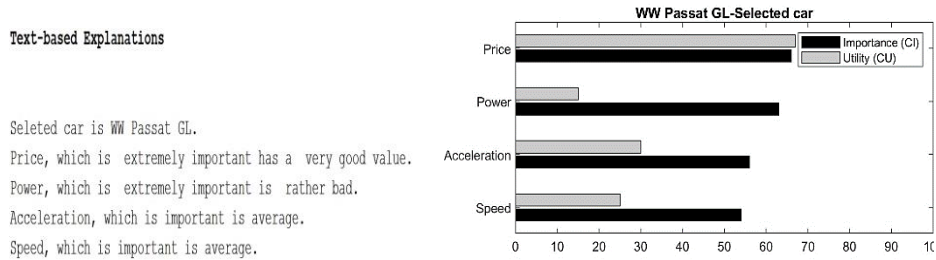

**Fig. 4.** Text-based and visual explanations for selected car WW Passat GL

The contrastive explanations are generated to compare the selected car example to other instances. Comparison between the selected car (WW Passat GL) to the expensive car (Citroen XM), and to the average car (WW Vento) is visually presented. Fig. 5 shows that the selected car has a very good value compare to the average car considering the importance of this criteria. Although the average car has higher utility values for acceleration and speed, it is exceeding the importance of the criteria (Selected car is better because of the importance and utility value of the Price criteria). Similarly, the expensive car has very low utility in terms of price, and it has quite high values for power, acceleration and speed compare to the selected car.

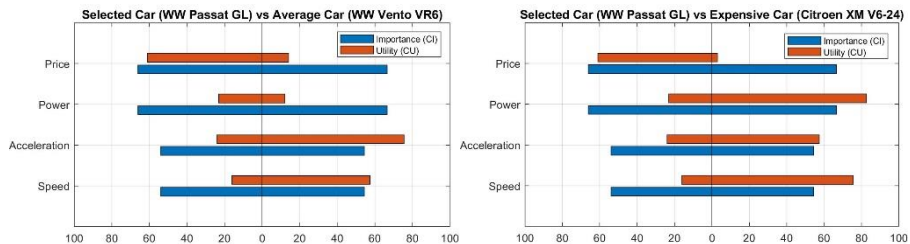

**Fig. 5.** Contrastive visual explanations for selected car (WW Passat GL) with average car (WW Vento VR6) and expensive car (Citroen XM V6-24)

### 5.2 Explaining Iris Flower Classification Using CI and CU Method

In this section, the results of explaining individual predictions using CI and CU on Iris flower classification is presented. The dataset contains 150 labeled flowers from the genus Iris. The trained network classifies Iris flowers into three species; Iris Setosa, Iris Versicolor and Iris Virginica based on the properties of leaves. These properties are namely; petal length, petal width, sepal length, sepal width. The trained network outputs the prediction value for each species which the highest one being the predicted class. The network is used to classify patterns that it has not seen before and results are used to generate explanations for individual instances.

An example of how explanations are generated based on CI and CU values is illustrated in Fig. 6. Given following input values; 7 (petal length), 3.2 (petal width), 6 (sepal length), 1.8 (sepal width), model predicts the class label as Iris Virginica. In order to compute the CI and CU values, we randomize 150 samples, and estimate $Cmax_x(C_i)$ and $Cmin_x(C_i)$ values for each input feature. The red-cross ($y_{i,j}$) indicates the current prediction value. Each figure demonstrates the importance of that feature and the usefulness for the predicted class. Similarly, CI and CU values of other classes are obtained to generate contrastive explanations. Note that a feature that is distinguishing for Iris Virginica may not be that distinguishing or important for other classes. The color bar indicates the contextual utility values converted into natural language expressions.

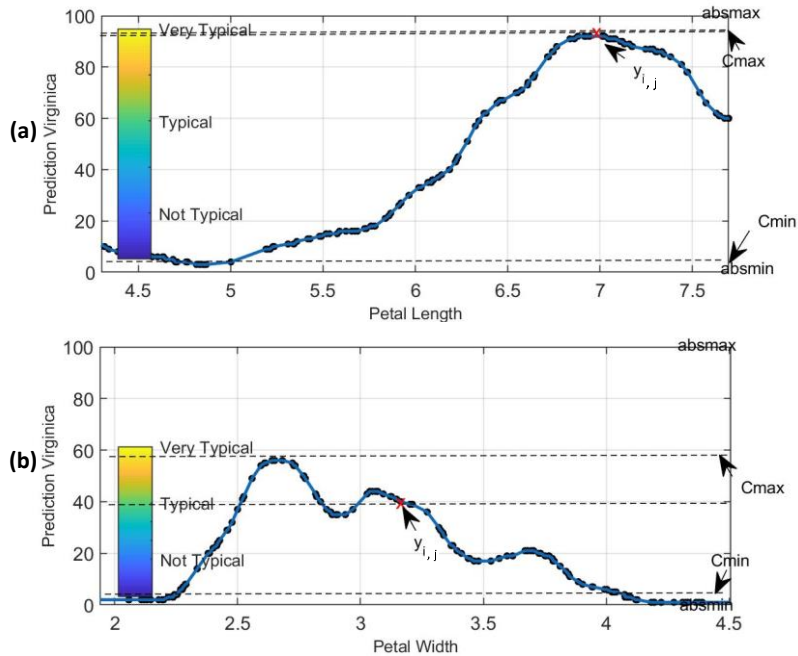

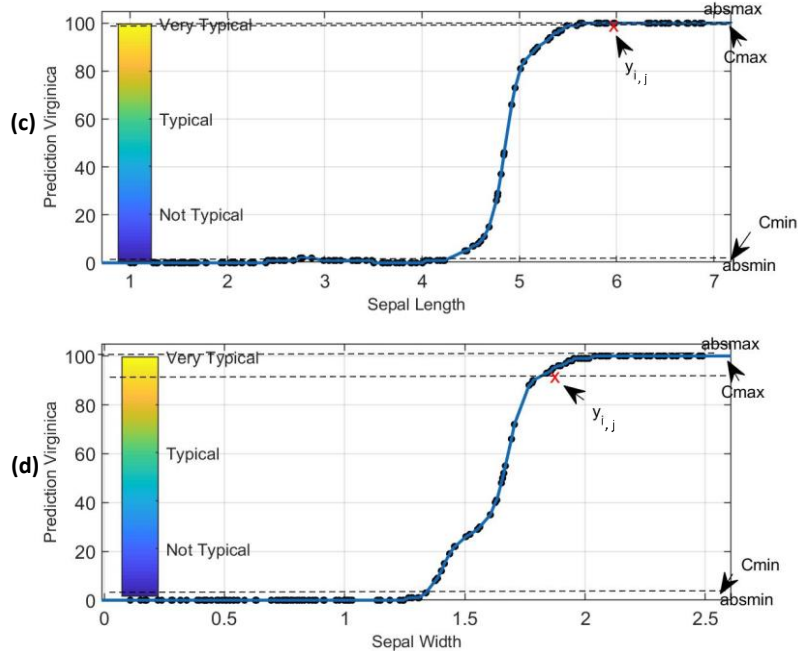

**Fig. 6.** CI and CU values of each features for Iris Virginica classification, **(a)** Petal length, **(b)** Petal width, **(c)** Sepal length, and **(d)** Sepal width

Table 2 shows the result of the sample test. For this case, sepal length is the most important feature with the highest utility value contributing to the class and the petal width is the least contributing feature for the given instance.

**Table 2.** $Cmin_x$, $Cmax_x$ and CIU values of each feature for the class label Iris Virginica

|      | **Petal Length** | **Petal Width** | **Sepal Length** | **Sepal Width** |
|------|------------------|-----------------|------------------|-----------------|
| **Cmin** | 3 | 1 | 0 | 0 |
| **Cmax** | 92 | 56 | 100 | 100 |
| **CI%** | 89 | 55 | 100 | 100 |
| **CU** | 1 $(y_{i,j}=.92)$ | 0.69 $(y_{i,j}=.39)$ | 1 $(y_{i,j}=1)$ | 0.91 $(y_{i,j}=.91)$ |

Table 3 shows how these values are transformed into natural language expressions to generate explanations based on the degree of the values.

**Table 3.** Symbolic representation of the CI and CU values

| **Degree($d$)** | **Contextual Importance** | **Contextual Utility** |
|-----------------|---------------------------|------------------------|
| $0 < d \leq 0.25$ | Not important | Not typical |
| $0.25 < d \leq 0.5$ | Important | Unlikely |
| $0.5 < d \leq 0.75$ | Rather important | Typical |
| $0.75 < d \leq 1.0$ | Highly important | Very typical |

The obtained values are translated into explanation phrases as shown in Fig. 7. These are the complete explanations which justifies why the model predicts this class label. Furthermore, the contrastive explanations are produced to demonstrate the contrasting cases. Fig. 8 shows the results of this application.

```
The model`s prediction is 98% Iris Virginica. Because;
The petal length  which is a highly important (CI=89%) feature has a very typical (CU=1) size.
The petal width  which is rather an important (CI=55%) feature has a typical (CU=0.69) size.
The sepal length  which is a highly important (CI=100%) feature has a very typical (CU=1) size.
The sepal width  which is a highly important (CI=100%) feature has a very typical (CU=0.91) size.
And the biggest contributing feature is the sepal length.
```

**Fig. 7.** Complete explanation for the class label Iris Virginica

```
It is not Iris Setosa(0%), because;
The petal length  which is a highly important (CI=86%) feature has not a typical (CU=0) size.
The petal width  which is a highly important (CI=98%) feature has an unlikely (CU=0.48) size.
The sepal length  which is a highly important (CI=100%) feature has not a typical (CU=0) size.
The sepal width  which is a highly important (CI=100%) feature has not a typical (CU=0) size.
It is not Iris Versicolor(2%), because;
The petal length  which is rather an important (CI=61%) feature has not a typical (CU=0.03) size.
The petal width  which is a highly important (CI=97%) feature has not a typical (CU=0.13) size.
The sepal length  which is a highly important (CI=99%) feature has not a typical (CU=0) size.
The sepal width  which is a highly important (CI=100%) feature has not a typical (CU=0.09) size.
```

**Fig. 8.** Contrastive explanations for Iris Setosa and Iris Versicolor

## 6 Discussion

Intelligent systems that are explaining their decisions to increase the user's trust and acceptance are widely studied. These studies propose various means to deliver explanations in form of; if-then rules [35], heat-maps [36], visuals [37], and human-labeled text [38]. These explanations and justifications provide limited representation of the cause of a decision. The CIU method presented here proposes two modalities as visuals and textual form to express relevant explanations. The variability in modality of presenting explanations could improve interaction quality, particularly in time-sensitive situations (e.g. switching to visual explanations from text-based explanations). Moreover, CI and CU values can be represented with different levels of details and produce explanations that are tailored to the users' specification. User-customized explanations could reduce ambiguity in reasoning. This is particularly important in safety-critical applications where users require a clear response from the system.

Explanation methods should be responsive to different types of queries. Most explanation methods only provide explanations which respond to why a certain decision or prediction was made. However, humans usually expect explanations with a contrasting case to place the explanation into a relevant context [8]. This study present examples of complete and contrastive explanation to justify the predicted outcomes. One stream of research propose justification based explanations for image dataset combining visual and textual information [39]. Although they produce convincing explanations for users, they offer post-hoc explanation which is generally constructed without following the model's reasoning path (unfaithfully).

Faithfulness to actual model is important to shows the agreement to the input-output mapping of the model. If the explanation method is not faithful to the original model then the validity of explanations might be questionable. While the rule extraction method produces faithful explanations, it is often hard to trace back the reasoning path, particularly when the number of features is too high. Other methods such as approximating an interpretable model provide only local fidelity for individual instances [29]. However, features that are locally important may not be important in the global context. CIU overcome the limitation of the above methods by providing explanations based on the highest and the lowest output values observed by varying the value of the input(s). However, accurate estimation of the minimal and the maximal values remains a matter of future studies. Furthermore, CIU is a model agnostic method which increases the generalizability of the explanation method in selection of the learning model.

## 7 Conclusion

The aim of this paper is presenting contextual importance and utility method to provide explanations for black-box model predictions. CIU values are represented as visuals and natural language expressions to increase the comprehensibility of the explanations. These values are computed for each class and features which enable to further produce contrastive explanation against the predicted class. We show the utilization of the CIU for linear and non-linear models to validate the generalizability of the method. Future work could extend the individual instance explanations to global model explanations in order to assess and select between alternative models. It is also valuable to focus on integrating CIU method into practical applications such as image labeling, recommender systems, and medical decision support systems. A future extension of our work relates to the CIU's utility in producing dynamic explanations by considering user's characteristics and investigating the usability of the explanations in real-world settings.

## Acknowledgment

This work was partially supported by the Wallenberg AI, Autonomous Systems and Software Program (WASP) funded by the Knut and Alice Wallenberg Foundation.

## References

1.      Biran, O. and Cotton C.: Explanation and justification in machine learning: A survey. In: IJCAI-17 Workshop on Explainable AI (XAI). (2017).

2. Anjomshoae, S., Najjar, A., Calvaresi, D., Främling, K.: Explainable Agents and Robots: Results from a Systematic Literature Review. In: Proceedings of the 18th International Conference on Autonomous Agents and Multiagent Systems (2019).
3. Samek, W., T. Wiegand, and Müller K. R.: Explainable artificial intelligence: Understanding, visualizing and interpreting deep learning models. arXiv preprint arXiv:1708.08296. ( 2017).
4. Ras, G., M. van Gerven, and Haselager P.: Explanation methods in deep learning: Users, values, concerns and challenges. In: Explainable and Interpretable Models in Computer Vision and Machine Learning. (2018), Springer. p. 19-36.
5. Nunes, I. and Jannach, D.: Interaction, A systematic review and taxonomy of explanations in decision support and recommender systems. In: User Modeling and User-Adapted Interaction, 27(3-5), 393-444. (2017).
6. Miller, T., Howe, P. and Sonenberg, L.: Explainable AI: Beware of inmates running the asylum or: How I learnt to stop worrying and love the social and behavioural sciences. arXiv preprint arXiv:1712.00547. (2017).
7. Främling, K.: Explaining results of neural networks by contextual importance and utility. In: Proceedings of the AISB'96 conference. (1996).
8. Miller, T.: Explanation in artificial intelligence: Insights from the social sciences. In: Artificial Intelligence. (2018).
9. Molnar, C.J.: Interpretable machine learning. A Guide for Making Black Box Models Explainable, In: Leanpub. (2018).
10. Shortliffe, E.: Computer-based medical consultations: MYCIN. Vol. 2.: Elsevier. (2012)
11. Clancey, W.J.: The epistemology of a rule-based expert system—a framework for explanation. In: Artificial intelligence, 20(3): p. 215-251. (1983).
12. Swartout, W.R.: Xplain: A system for creating and explaining expert consulting programs. In: University of Southern California Marina Del Rey Information Sciences Inst. (1983)
13. Swartout, W., Paris, C. and Moore, J.: Explanations in knowledge systems: Design for explainable expert systems. In: IEEE Expert 6, no. 3: 58-64. (1991).
14. Langlotz, C., Shortliffe, E.H. and Fagan, L.M.: Using Decision Theory to Justify Heuristics. In: AAAI. (1986).
15. Klein, D.A.: Decision-analytic intelligent systems: automated explanation and knowledge acquisition. Routledge, (2013).
16. Forsythe, D.E.: Using ethnography in the design of an explanation system. In: Expert systems with Applications. 8(4): p. 403-417. (1995).
17. Henrion, M. and Druzdzel, M.J.: Qualtitative propagation and scenario-based scheme for exploiting probabilistic reasoning. In: Proceedings of the Sixth Annual Conference on Uncertainty in Artificial Intelligence. Elsevier Science Inc. (1990).
18. Lacave, C. and Díez, F.J.: A review of explanation methods for Bayesian networks. 17(2): p. 107-127. (2002).
19. Nunes, I. and Jannach, D.: A systematic review and taxonomy of explanations in decision support and recommender systems. In: User Modeling and User-Adapted Interaction, 27(3-5): p. 393-444. (2017).

20. Souillard-Mandar, W., Davis, R., Rudin, C., Au, R. and Penney, D.: Interpretable Machine Learning Models for the Digital Clock Drawing Test. arXiv preprint arXiv:1606.07163. (2016).
21. Ustun, B. and Rudin, C.: Supersparse linear integer models for optimized medical scoring systems. In: Machine Learning, 102(3): p. 349-391. (2016).
22. Lakkaraju, H., Bach, S.H. and Leskovec, J.: Interpretable decision sets: A joint framework for description and prediction. In: Proceedings of the 22nd ACM SIGKDD international conference on knowledge discovery and data mining. (2016).
23. Rudin, C., Letham, B. and Madigan, D.: Learning theory analysis for association rules and sequential event prediction. In: The Journal of Machine Learning Research 14(1): p. 3441-3492. (2013).
24. Letham, B., Rudin, C., McCormick, T.H. and Madigan, D.: Building interpretable classifiers with rules using Bayesian analysis. In: Department of Statistics Technical Report tr609, University of Washington (2012).
25. Kim, B., Rudin, C. and Shah, J.A.: The bayesian case model: A generative approach for case-based reasoning and prototype classification. In: Advances in Neural Information Processing Systems. (2014).
26. Kim, B., Shah, J. A., & Doshi-Velez, F.: Mind the gap: A generative approach to interpretable feature selection and extraction. In: Advances in Neural Information Processing Systems. (2015).
27. Kononenko, I.: An efficient explanation of individual classifications using game theory. In: Journal of Machine Learning Research, 11(Jan): p. 1-18. (2010).
28. Krause, J., Perer, A. and Ng, K.: Interacting with predictions: Visual inspection of black-box machine learning models. In: Proceedings of the 2016 CHI Conference on Human Factors in Computing Systems. (2016).
29. Ribeiro, M.T., Singh, S. and Guestrin, C.: "Why Should I Trust You?", In: Proceedings of the 22nd ACM SIGKDD International Conference on Knowledge Discovery and Data Mining - KDD '16. p. 1135-1144. (2016).
30. Ribeiro, M.T., Singh, S. and Guestrin, C.: Model-agnostic interpretability of machine learning. arXiv preprint arXiv:1606.05386 (2016).
31. Främling, K.: Modélisation et apprentissage des préférences par réseaux de neurones pour l'aide à la décision multicritère. Institut National de Sciences Appliquées de Lyon, Ecole Nationale Supérieure des Mines de Saint-Etienne, France, (1996). 209 p.
32. Främling, K. and Graillot D.: Extracting Explanations from Neural Networks. In: Proceedings of the ICANN. (1995).
33. Saaty, T.L.: Decision making for leaders: the analytic hierarchy process for decisions in a complex world. RWS publications. (1990).
34. Roy, B.: Classement et choix en présence de points de vue multiples. 2(8): p. 57-75. (1968).
35. Setiono, R., Azcarraga, A. and Hayashi, Y.: MofN Rule Extraction from Neural Networks Trained with Augmented Discretized Input. In: Proceedings of the 2014 International Joint Conference on Neural Networks (Ijcnn), (2014): p. 1079-1086.
36. Seo, D., Oh, K. and Oh, I.S.: Regional Multi-scale Approach for Visually Pleasing Explanations of Deep Neural Networks. arXiv preprint arXiv:1807.11720 (2018).

37. Štrumbelj, E. and Kononenko, I.: A General Method for Visualizing and Explaining Black-Box Regression Models. In: Adaptive and Natural Computing Algorithms, Pt Ii, (2011). 6594: p. 21-30.
38. Berg, T. and Belhumeur, P.N.: How Do You Tell a Blackbird from a Crow?. In: IEEE International Conference on Computer Vision. p. 9-16. (2013).
39. Huk Park, D., Anne Hendricks, L., Akata, Z., Rohrbach, A., Schiele, B., Darrell, T. and Rohrbach, M.: Multimodal explanations: Justifying decisions and pointing to the evidence. In: Proceedings of the IEEE Conference on Computer Vision and Pattern Recognition. (2018).